\documentclass[letterpaper, 10 pt, journal, twoside]{IEEEtran}

\IEEEoverridecommandlockouts                              

\usepackage{graphicx}
\usepackage{amsmath} 
\usepackage{enumitem}
\usepackage{siunitx}
\usepackage{textcomp}
\usepackage{multirow}

\title{
An Origami-Inspired Variable Friction Surface for Increasing the Dexterity of Robotic Grippers
}

\author{Qiujie Lu, \IEEEmembership{Student Member, IEEE}, Angus B. Clark, \IEEEmembership{Student Member, IEEE}, Matthew Shen, and Nicolas Rojas, \IEEEmembership{Member, IEEE}%
\thanks{Manuscript received: September, 10th, 2019; Revised December, 31st, 2019; Accepted January, 26th, 2020.}
\thanks{This paper was recommended for publication by Editor Hong Liu upon evaluation of the Associate Editor and Reviewers' comments. This work was supported in part by the Engineering and Physical Sciences Research Council grant EP/R020833/1.} 
\thanks{Qiujie Lu, Angus B. Clark, Matthew Shen, and Nicolas Rojas are with the REDS Lab, Dyson School of Design Engineering, Imperial College London, 25 Exhibition Road, London SW7 2DB, UK.  (e-mail: {\tt\footnotesize
 \{q.lu17, a.clark17, matthew.shen17, n.rojas\}@imperial.ac.uk})}%
\thanks{Digital Object Identifier (DOI): see top of this page.}
}

\begin{document}

\maketitle

\begin{abstract}
While the grasping capability of robotic grippers has shown significant development, the ability to manipulate objects within the hand is still limited. One explanation for this limitation is the lack of controlled contact variation between the grasped object and the gripper. For instance, human hands have the ability to firmly grip object surfaces, as well as slide over object faces, an aspect that aids the enhanced manipulation of objects within the hand without losing contact. In this letter, we present a parametric, origami-inspired thin surface capable of transitioning between a high friction and a low friction state, suitable for implementation as an epidermis in robotic fingers. A numerical analysis of the proposed surface based on its design parameters, force analysis, and performance in in-hand manipulation tasks is presented. Through the development of a simple two-fingered two-degree-of-freedom gripper utilizing the proposed variable-friction surfaces with different parameters, we experimentally demonstrate the improved manipulation capabilities of the hand when compared to the same gripper without changeable friction. Results show that the pattern density and valley gap are the main parameters that effect the in-hand manipulation performance. The origami-inspired thin surface with a higher pattern density generated a smaller valley gap and smaller height change, producing a more stable improvement of the manipulation capabilities of the hand.
\end{abstract}


\section{Introduction}
\IEEEPARstart{D}{ue} to their mechanical simplicity, low cost, reliability, and low control complexity, two-fingered low-degree-of-freedom robot grippers are prevalent in industrial tasks, especially for pick and place operations \cite{guo2017design}. However, robotic research has long been interested in not only the ability to grasp, but also the in-hand manipulation of a varied set of objects in order to improve the dexterity and applicability of robots. Many efforts have indeed been made to replicate the functionality of the human hand, the best example of a dexterous system. Well controlled anthropomorphic robotic hands have been developed \cite{ueda2010multifingered, chella2004posture, shi2017dynamic}, which are good at performing complex hand gestures. However, these systems have shown significant limitations and challenges for in-hand manipulation in unstructured environments due to over-constrained structures, uncertainties, and compound errors and failures in actuation and sensing. Improving the in-hand manipulation ability of robot grippers without increasing their design and control complexity has since become an active area of research in recent years \cite{rojas2016gr2, bircher2017two, dafle2014extrinsic, chavan2015two, chavan2018pneumatic}.

\begin{figure}[t!]
    \centering
    \includegraphics[width=\columnwidth]{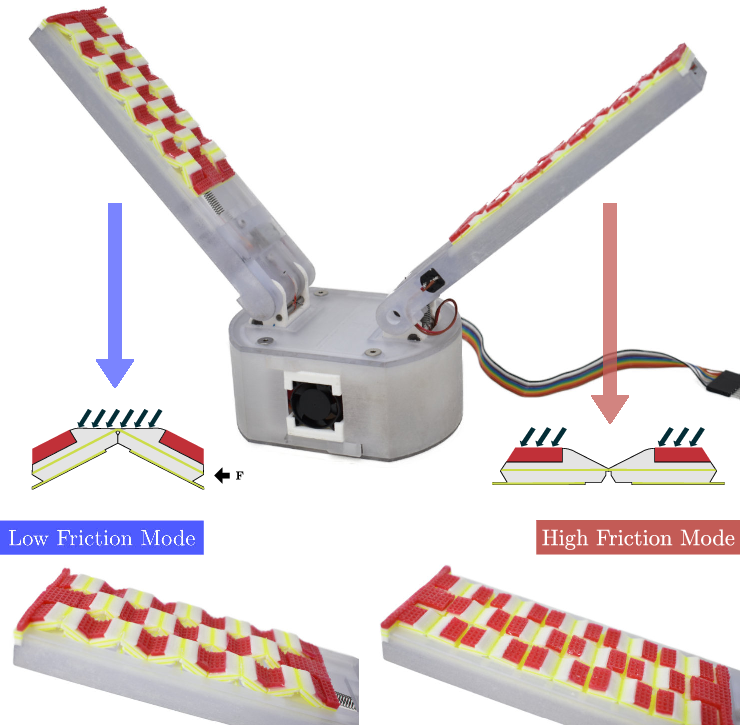}
    \caption{Two-fingered two-degree-of-freedom gripper with fingers using the proposed origami-inspired variable friction (O-VF) surface. The controlled states of low friction (left finger) and high friction (right finger) are depicted, demonstrating the varying contact surfaces (black arrows).}
    \label{mainPrototype}
\end{figure}

The most common in-hand manipulations for robot grippers that have been studied are sliding and rotating operations. Chavan-Dafle \textit{et al.} achieved spinning point contact and firm contact by changing the finger-object contact geometry and varying the gripping force \cite{chavan2015two,chavan2018pneumatic}. Objects reorient about the axis between the contact points from a horizontal pose to a vertical pose due to gravity, however with limited reorientation direction and range. In-hand reorientation of grasped objects has been also demonstrated using tactile feedback \cite{ward2017model}. Alternatively, adding or changing components of existing hand mechanisms is a common method for improving robot gripper abilities. The GR2 gripper increased the object range of motion by introducing an elastic pivot joint between the two fingers \cite{rojas2016gr2}, and Terasaki \textit{et al.} designed a rotation mechanism attached to the tips of a parallel two-fingered gripper combined with a motion planning system to increase dexterity \cite{terasaki1994motion}. 

In human hands, the soft and pulpy tissue of the fingertip can comply around the shape of objects, gripping them firmly when a force is applied. The use of soft material on gripper fingertips provides a compromise between compliance and strength \cite{chorley2008biologically}, and has been shown to provide a larger workspace and adaptability \cite{lu2019soft}. With a more rigid, smooth contact however, sliding can be achieved more easily as the system behaves as an inclined plane. In fact, friction also plays an important role in object manipulation \cite{tomlinson2007review}. The frictional properties of biological skin has been investigated to show that the effects of these parameters are essential for feedback and forward gripping control systems. Comaish and Bottoms for instance showed that the coefficient of friction between the skin and various materials is not portrayed by the simple laws of friction, but by a complex viscoelastic relationship, especially under hydrated and lubricated environments \cite{comaish1971skin}; while Adams \textit{et al.} concluded that the human finger pad contact frictions are complex, and mainly influenced by fingerprint ridges \cite{adams2013finger}. By assuming the surfaces of the fingerprint ridges as thin water films, these last authors observed a decrease in friction at larger sliding velocities.

\begin{figure}[t!]
    \centering
    \includegraphics[width=\columnwidth]{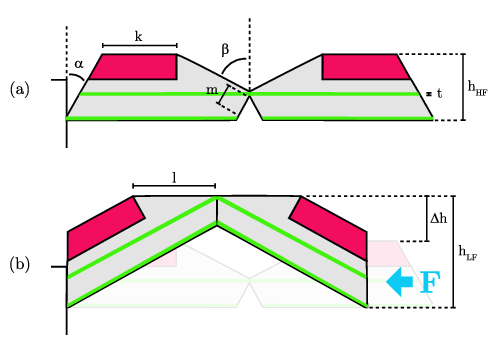}
    \caption{Specifications of the folding pattern, defining the area ratio of variable friction surfaces and change in thickness between modes: \textbf{(a)} high friction and \textbf{(b)} low friction.}
    \label{foldingSpecificationsSymbols}
\end{figure}

Spiers \textit{et al.} described a variable friction principle analogy to the human finger pad, and presented a passive and active variable friction robot finger design to achieve a similar effect \cite{spiers2018variable}. They discussed that human finger pads can perform sliding via light contact on the epidermal layer and pivoting via heavier touch with compression of glabrous fat. This behaviour was then emulated via a suspended low friction surface, where an object can slide on a low friction surface and rotate on high friction surface. This variable friction finger design had the ability to change the friction mode and achieve isolated translation and rotation using a simple two-fingered two-degree-of-freedom gripper. Following this principle, in this letter we propose a novel origami-inspired thin surface for robotic fingers which allows obtaining the benefits of variable friction for dexterity in a much more compact setting.

The introduced origami-inspired variable friction (O-VF) surface, based on a deformation-limited mountain/valley fold structure, exposes two different contact surfaces (materials) using a single on-off actuator. Fig.~\ref{mainPrototype} shows the proposed concept and a two-fingered two-degree-of-freedom gripper with fingers equipped with O-VF surfaces. Thanks to the possibility of controlling states of low and high contact friction, the fingers of the robot hand can either slide over objects or firmly grasp them, similar to a human finger, without increasing the complexity of the control problem significantly. The rest of the paper is as follows: In Section II, we detail the design of the novel origami-inspired variable friction surface, and analyze the design parameters and required folding force. We then present the implementation of multiple prototypes with various design parameters, and evaluate the in-hand manipulation (translation and rotation) performance of the developed O-VF surfaces with objects of different size and shape (section III). Finally, we discuss the experiment results in section IV, and conclude in section V.

\section{Design and Numerical Analysis of O-VF Surface}

\begin{figure}[t!]
    \centering
    \includegraphics[width=\columnwidth]{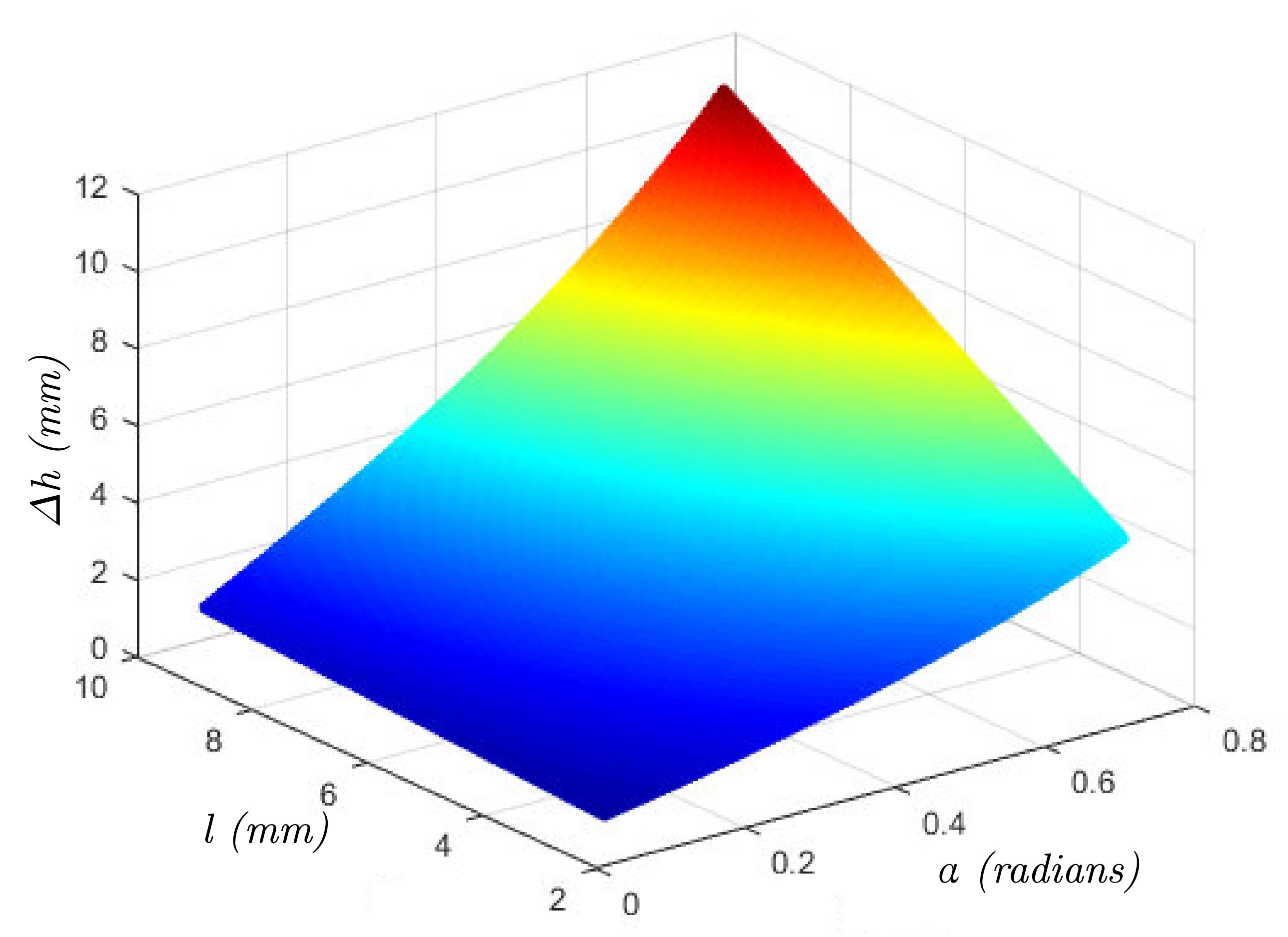}
    \caption{Surface plot showing the relationship between the length of low friction area (l), the folding angle ($\alpha$), and change in thickness of the overall structure between friction modes ($\Delta$h).}
    \label{symbolRelationship}
\end{figure}

Using origami folding processes, complex robots can be fabricated by simple approaches \cite{rus2018design}. Due to the diversity of origami patterns, these folding processes create a large number of potential possibilities. For instance, the Mirura-Ori pattern allows the entire structure to be folded or unfolded in two directions \cite{miura1985method} using a single motion. Alternatively the Kresling Crease pattern, which resembles a chiral tower, combines longitudinal and rotational motion simultaneously, similar to a screw motion \cite{kresling2012origami}. Our surface design is based on a deformation-limited accordion pattern, which allows for the changing of friction modes using only one actuator. The detailed design process of this structure is shown next.

\renewcommand{\arraystretch}{1.1}
\begin{table*}
\centering
\caption{Coefficient of friction between ABS and different materials.\protect\\PETG, ABS, and PLA are sanded to obtain a smooth surface, whereas the Ecoflex are tested with surface finishes: planar, ridged, and checkered.}
\label{friction}
\begin{tabular}{ccccccccccccc} 
\hline
\multirow{2}{*}{\textbf{Material}} & \multirow{2}{*}{\textbf{PETG}} & \multirow{2}{*}{\textbf{ABS}} & \multirow{2}{*}{\textbf{PLA}} & \multicolumn{3}{c}{\textbf{Ecoflex 00-10}} & \multicolumn{3}{c}{\textbf{Ecoflex 00-20}} & \multicolumn{3}{c}{\textbf{Ecoflex 00-30}}\\
                          &                       &                      &                      & \includegraphics[width=0.8cm]{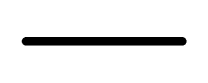} & \includegraphics[width=0.8cm]{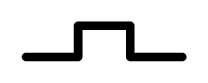} & \includegraphics[width=0.8cm]{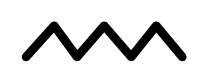} & \includegraphics[width=0.8cm]{Figures/table-p.eps} & \includegraphics[width=0.8cm]{Figures/table-r.eps} & \includegraphics[width=0.8cm]{Figures/table-c.eps} & \includegraphics[width=0.8cm]{Figures/table-p.eps} & \includegraphics[width=0.8cm]{Figures/table-r.eps} & \includegraphics[width=0.8cm]{Figures/table-c.eps} \\
\cline{2-13}
\textbf{Coefficient of friction ($\mu$)} & 0.08 & 0.08 & 0.08 & 0.63  & 0.72 & 0.77 & 0.57 & 0.55 & 0.64 & 0.52 & 0.54 & 0.61 \\ 
\hline
\end{tabular}
\end{table*}

To allow for a variation in friction, and due to the lack of single materials with easily variable friction, the finger contact surface had to contain both a low friction component and a high friction component. By altering the configuration of the surface structure, the exposed component would act as the current overall friction of the surface. To allow for this, a deformation-limited accordion fold structure was proposed, where an angular change of the surface raised the active friction component, whilst preventing contact with the alternate friction component. The working principle of this design can be seen in Fig. \ref{foldingSpecificationsSymbols}, where under the compression of a force the structure folds up to a pre-defined limit, changing the outer-most face in contact with a grasped object.

\subsection{Parametric design}
The folding structure was defined in a parametric form, allowing for the variation of the structure based on the desire of the user. The variables that define the topology, detailed in Fig.~\ref{foldingSpecificationsSymbols}, are the length of the low friction area, $l$, and the length of the high friction area, $k$, which we define as a percentage of $l$ as $k = R\,l$. We also define the folding angle, $\alpha$, as the bending angle required to transition between friction modes. To ensure a flat surface is achieved after folding, we also define the low friction offset angle, $\beta$, such that $90^\circ = \alpha + \beta$.

The thickness of the folding layers of the structure, $t$, depends on the strength and flexibility of the selected material. To prevent the structure from over-folding, we implemented limiting faces, defined by length $m$, as indicated in Fig.~\ref{foldingSpecificationsSymbols}, which prevents the structure from excessive folding due to their contact once folded. Fig.~\ref{foldingSpecificationsSymbols} shows a single folding unit of the surface, where the number of units in one surface (the pattern density) is represented by $N$. The change in thickness of the overall structure between the friction modes, $\Delta h$, is defined as $\Delta h = h_{LF} - h_{HF}$, where $h_{HF}$ is the thickness in high friction mode and $h_{LF}$ is the thickness in low friction mode. It can be verified via trigonometry that:
\begin{align*}
h_{LF} &=  (Rl\sin{\alpha}+\frac{h_{HF}}{\cos{\alpha}})\textrm{ and} \\
h_{HF} &= 2t + l\sin{\alpha} + m\cos{\alpha}.
\end{align*}

\begin{figure}[t!]
    \centering
    \includegraphics[width=\columnwidth]{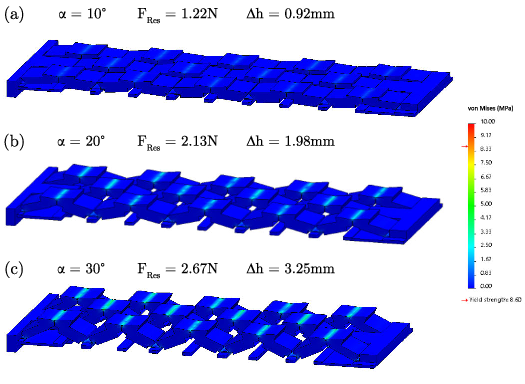}
    \caption{Static simulation of the deformation required to fully fold each design with $\alpha$ at values \textbf{(a)} 10\textdegree , \textbf{(b)} 20\textdegree , and \textbf{(c)} 30\textdegree . The resultant force (N) required to fully fold each specification is also shown. Simulation surface colours indicate the observed stress on the thermoplastic polyurethane (TPU) material.}
    \label{foldingSimulation}
\end{figure}

We can see the $\Delta h$ is related to $\alpha$, $l$, $R$, $t$, and $m$, where the minimum value of the thickness of the folding layers $t$ is dependent on the 3D printer resolution and the minimum limiting face $m$ size required to prevent over bending. For our design, we considered these two parameters as constants and selected values that give a reduced thickness and a reliable performance with $t$ = 0.3 mm, and $m$ = 2 mm. We analyzed the relationship between the rest of the parameters and the change in thickness of the overall structure between friction modes, with results shown in Fig.~\ref{symbolRelationship}. We applied further constraints on the parameters due to the manufacturing capability and the entire surface length, resulting in $\alpha$ set between 0.175 and 0.785 radians and $l$ between 3 and 10 mm. From Fig.~\ref{symbolRelationship}, we can see that a smaller value for $\alpha$ and $l$ give the minimum change in overall height between friction modes.

\subsection{Material selection}
Knowledge of the stress experienced when folded was required to ensure the elastic limit of the material composing the joints was not exceeded, causing permanent deformation of the structure, thus allowing the structure to revert to its previous unfolded state when deactivated. To calculate this stress, finite element analysis was performed through Solidworks Simulation for both the material Acrylonitrile Butadiene Styrene (ABS) and Thermoplastic Polyurethane (TPU), with rough yield strengths of 39 MPa and 8.6 MPa, respectively \cite{ultmat}, as the design was intended to be easily 3D printed by a common, single nozzle, desktop 3D printer. The results showed a maximum von Mises stress of 540 MPa for ABS, and 4.6 MPa for TPU. Had the structure been printed out solely of the significantly more rigid ABS, it would have therefore undergone plastic deformation. To prevent this, while ensuring the rest of the structure remains rigid, layers of TPU (shown in green in Fig. \ref{foldingSpecificationsSymbols}) were introduced to the model at the stress concentrations at the folding areas, which as indicated by the simulation would not exceed the elastic limit. The simulation results for the experienced stress of the structure formed from TPU can be seen in Fig.~\ref{foldingSimulation}, with the yield stress indicated on the colour legend.

Fig.~\ref{foldingSimulation} also presents the force required to fold the structure, with $\alpha$ at values 10\textdegree, 20\textdegree, and 30\textdegree. As $\alpha$ increases, the distance over which a force needed to be applied increased, with a 1.9 mm compression at 10\textdegree, 7.5 mm compression at 20\textdegree, and 16.8 mm compression at 30\textdegree. The resultant force needed for compression also increased with $\alpha$, with a maximum force of 2.67N at $\alpha$=30\textdegree, as shown in Fig.~\ref{foldingSimulation}. In selecting a value for $\alpha$, $\Delta$h had to also be considered. For low values of $\alpha$ a low $\Delta$h was produced, which is ideal for minimising the height of the overall structure. However, $\Delta$h must also be large enough to ensure no accidental contact is made with the alternate friction surface. Therefore, a value of 30\textdegree\ was selected for $\alpha$ to ensure no unwanted contact, and the compression force was deemed respectable at small values across all values of $\alpha$.

\begin{figure}[t!]
    \centering
    \includegraphics[width=\columnwidth]{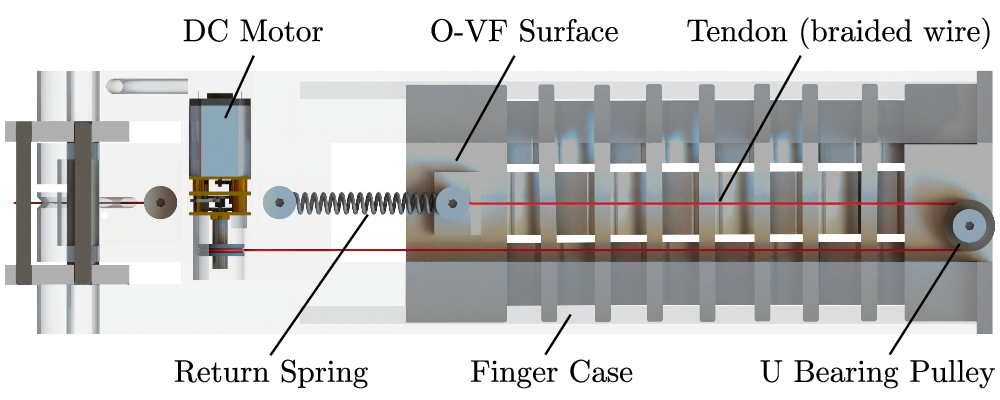}
    \caption{Section view of the CAD model finger showing the actuation method and tendon routing on the rear of the O-VF surface.}
    \label{CADBack}
\end{figure}

In the selection of materials for the folding structure friction surfaces, the coefficient of friction between ABS and 6 types of materials in 12 conditions were measured, shown in Table \ref{friction}. ABS was used as a constant comparative surface, as it allowed multiple objects to be easily 3D printed later for manipulation. For the low friction material, ABS, Polylactic Acid (PLA), and Polyethylene Terephthalate Glycol (PETG) were tested after sanding the raw 3D printed surface. The testing samples gave identical results, therefore ABS was chosen as it formed the strongest bond (and thus printed the best) with TPU. For the high friction material, three types of silicone with varying hardness (SmoothOn Ecoflex 00-10, 00-20, and 00-30) were tested in 3 conditions: Planar (indicated by a straight line), ridged (indicated by a line with a 'step'), and checkered (indicated by a zig-zag line). The highest coefficient of friction was shown by Ecoflex 00-10 in a checkered pattern, which was therefore selected for the high friction surface.

\section{Prototype Implementation}
\subsection{Prototype Design}
The O-VF surface was designed, as mentioned, to be easily 3D printed by a standard, desktop 3D printer with single nozzle. This allowed the low friction and high friction contact surface to be customized by changing the printing materials, or by adding a new material using layer deposition. In this work, the low friction contact surface is formed from ABS material and the high friction contact surface from Eco-Flex 10 silicone moulded with a checkered pattern. In our fabrication process, we first printed the O-VF surface structure, shown in Fig. \ref{foldingSimulation}, in ABS material for rigid parts with TPU for flexible hinges using a CraftBot Plus printer. As most consumer 3D printers are not capable of printing silicone, the silicone surface was molded separately and attached to the surface using adhesive. Depending on the material selection, it is straightforward for the fabrication process of the O-VF surface to be simplified to a single step using a multi-material 3D printer.

The selected actuation method for the O-VF surface was tendon driven via DC geared motors. A DC geared motor is slotted into the finger and connected with the O-VF surface via braided wire (SeaKnight 15 lb Classic Line). A 10 mm diameter U bearing pulley is attached at the back of the surface for smooth actuation and provides a central activation force. In the initial flat position, the surface is in high friction mode. By activating the DC motor, the surface is folded into the low friction mode. Once folded, the motor is deactivated, and the resistance of the DC motor gearbox maintains the low friction mode. When the DC geared motor is reversed, the structure is pulled back to its initial flat configuration by a tension spring. The arrangement of these components in the rear of each finger can be seen in Fig.~\ref{CADBack}.

\begin{figure}[t!]
    \centering
    \includegraphics[width=\columnwidth]{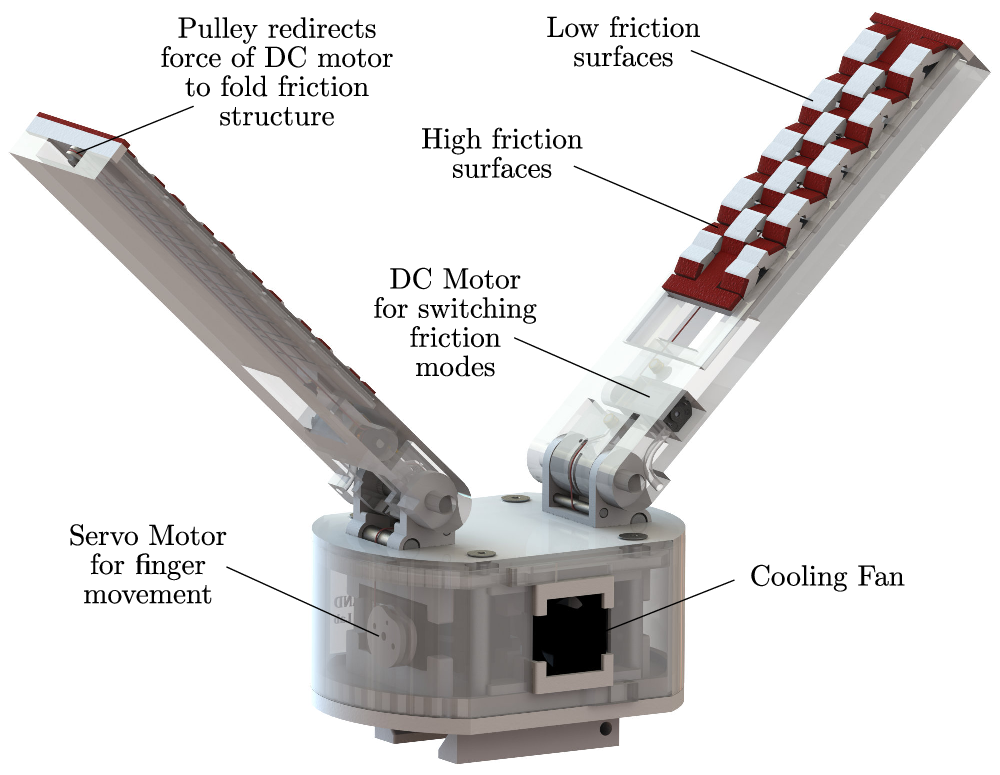}
    \caption{CAD model profile of the developed gripper showing the visual and positional difference of the variable stiffness surfaces (left finger in high-friction mode, right finger in low-friction mode), as well as the positioning of the motors for finger motion and control of the friction.}
    \label{prototypeProfile}
\end{figure}

\subsection{Experimental Setup}
\begin{figure}[t!]
    \centering
    \includegraphics[width=\columnwidth]{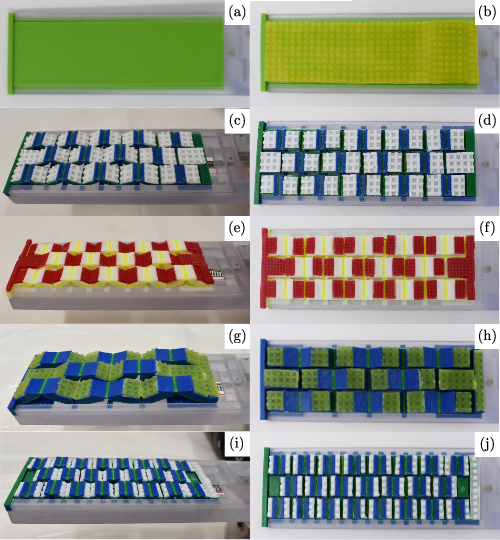}
    \caption{Testing surfaces: \textbf{(a)} low friction normal surface, \textbf{(b)} high friction normal surface, \textbf{(c, d)} weighted O-VF surface, \textbf{(e, f)} O-VF surface with medium density, \textbf{(g, h)} O-VF surface with low density, and \textbf{(i, j)} O-VF surface with high density.}
    \label{Fingersurface}
\end{figure}

The developed two-fingered two-degree-of-freedom gripper (Fig.~\ref{prototypeProfile}) was based upon the design of the Yale OpenHand \cite{ma2017yale}, with several modifications. The fingers are redesigned to contain two slots to accommodate the O-VF surfaces and DC geared motors (dimension of 173 x 51 x 16.3 mm), and are controlled by tendons driven by two Power HD servo motors (LF-20MG). All of the hand components were printed on Stratasys Objet 260/500 printers and CraftBot Plus printer in Vero Clear and ABS, respectively. The weight of the hand averages around 650 g (varying with different O-VF surfaces) and the dimension of the hand base is 109 x 88 x 55 mm.

Six different finger surfaces were developed to evaluate the design parameters' effect on the developed gripper. Fig.~\ref{Fingersurface} shows the appearances of the finger surfaces. Type (a) and (b) are typical finger surfaces with low friction (ABS) and high friction (EcoFlex-10), respectively. Type (c), (e), (g), and (i) are all surfaces based on the O-VF folding structure, representing a weighted surface with differing values for $k$ and $l$ (c), a medium density surface where $N$ = 5 (e), a low density surface where $N$ = 3 (g), and a high density surface where $N$ = 8 (i). Based on the previous parametric analysis, some parameters remain constant with their optimum value across all O-VF surfaces, where $\alpha$ = 30\textdegree, $m$  = 2 mm, and $t$ = 0.3 mm. For the surface (c) with weighted $l$ and $k$ length, where $l$ = 3 mm and $k$ = 8 mm, the $\Delta$h = 4.59 mm. The other three O-VF surfaces with middle ($l=k=5$ mm), low ($l=k=9.6$ mm), and high ($l=k=4$ mm) unit density, produce a $\Delta$h equal to 3.25 mm, 5.90 mm, and 2.67 mm, respectively.

The hand and O-VF surface were controlled by an Arduino Mega. A simple control approach was applied, using position control for the servo motors and time and speed control for the DC geared motors. To achieve this control of the DC geared motors, a H bridge was used providing both speed and direction control, with time control provided by the Arduino.

\begin{figure}[t!]
    \centering
    \includegraphics[width=\columnwidth]{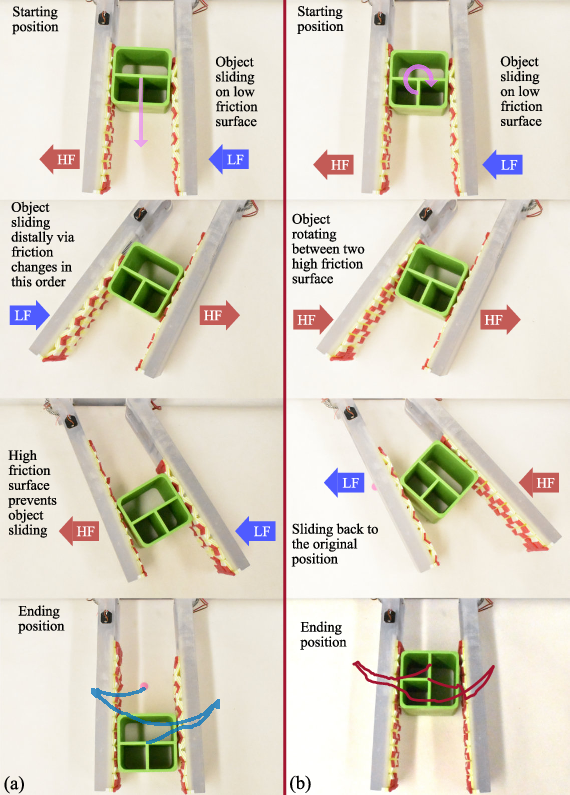}
    \caption{Method for achieving translation \textbf{(a)} and rotation \textbf{(b)} of a 50 mm width square object by actively controlling the variable friction surfaces while actuating the fingers. HF indicates high friction surface activated. LF indicates low friction surface activated. The arrow represents the finger moving direction. The blue and red curves in the final row show the trajectories of the manipulated objects. Further object manipulations can be seen in the accompanying video.}
    \label{figure}
\end{figure}

\subsection{In-hand Manipulation Realization}\label{sec:procedures}

From inspiration in how humans can manipulate an object in hand, we implemented control sequences for both rotating and translating an object between two fingers by applying different frictions while rotating the fingers.

Active finger object translation: Fig.~\ref{figure}(a) illustrates the distal translation of a 50 mm square from the base of the phalanges to the fingertips of both fingers using the developed gripper with O-VF surfaces. The control sequence for this distal translation is as follows:

\begin{enumerate}[]
    \item Actuate two fingers to grasp the object in the starting position. Switch to low friction mode on right finger.
    \item Rotate the fingers clockwise to maximum angle. The object slides distally on the right finger. When the fingers reach to the maximum angle, switch both finger surface friction modes (left to low, right to high).
    \item Rotate the fingers anticlockwise to maximum angle. The object slides distally on the left finger. When the fingers reach to the maximum angle, switch both finger surface friction modes again (left to high, right to low).
    \item Rotate the fingers clockwise back to the starting position.
\end{enumerate}

Active finger object rotation: Fig. \ref{figure}(b) illustrates how to actively rotate and translate a 50 mm square to achieve an isolated rotation (only rotation with no translation compared to the starting position). The control sequences for this isolated rotation are as follows:

\begin{enumerate}[]
    \item Actuate two fingers to grasp the object in the starting position. Switch to low friction mode on right finger. Rotate the fingers clockwise to let the object slide distally on the right finger to prepare a larger rotation interspace. 
    \item Rotate the fingers anticlockwise with both fingers in high friction surface mode to maximum angle. The object rotates in the fingers by pivoting on a corner. When the fingers reach the maximum angle, switch left finger surface friction mode to low. 
    \item Rotate the fingers clockwise back to the starting position.
\end{enumerate}

Both sequences can be repeated to achieve larger movement, and are suitable for objects in different size and shape. By reversing the methodology, translation and rotation in the opposite direction can be achieved (proximal translation towards the base of the fingers, counter clockwise rotation).

\begin{figure}[t!]
    \centering
    \includegraphics[width=\columnwidth]{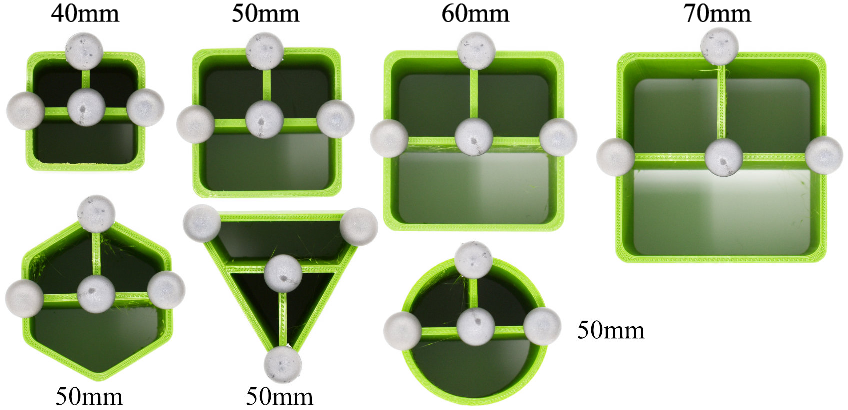}
    \caption{The 7 different objects evaluated: 4 squares of width 40 mm to 70 mm and 3 alternative shapes of width 50 mm. The height of all the objects is 60 mm. Motion tracking marker positions for each shape are also shown.}
    \label{objects}
\end{figure}

\section{Performance Evaluation}
In this section, we evaluate the translational and rotational performance of the 6 types of friction surfaces on the designed gripper using 7 different objects (4 squares of width 40 mm to 70 mm and 3 alternative shapes of width 50 mm) detailed in Fig.~\ref{objects}. The gripper was mounted on a Universal Robots UR5 robot arm with the objects placed on a planar surface, and followed the procedures described in section \ref{sec:procedures}. The control algorithm of the servo motor was kept consistent while testing all 6 finger surfaces, and each finger rotates to its maximum position to cover the full gripper workspace. The trajectories of the testing objects were recorded by motion tracking cameras (OptiTrack Flex3) and the results were post processed in MATLAB. Each test consisted of 5 repeated trials to generate reliable performance results.

\begin{figure}[t!]
    \centering
    \includegraphics[width=\columnwidth]{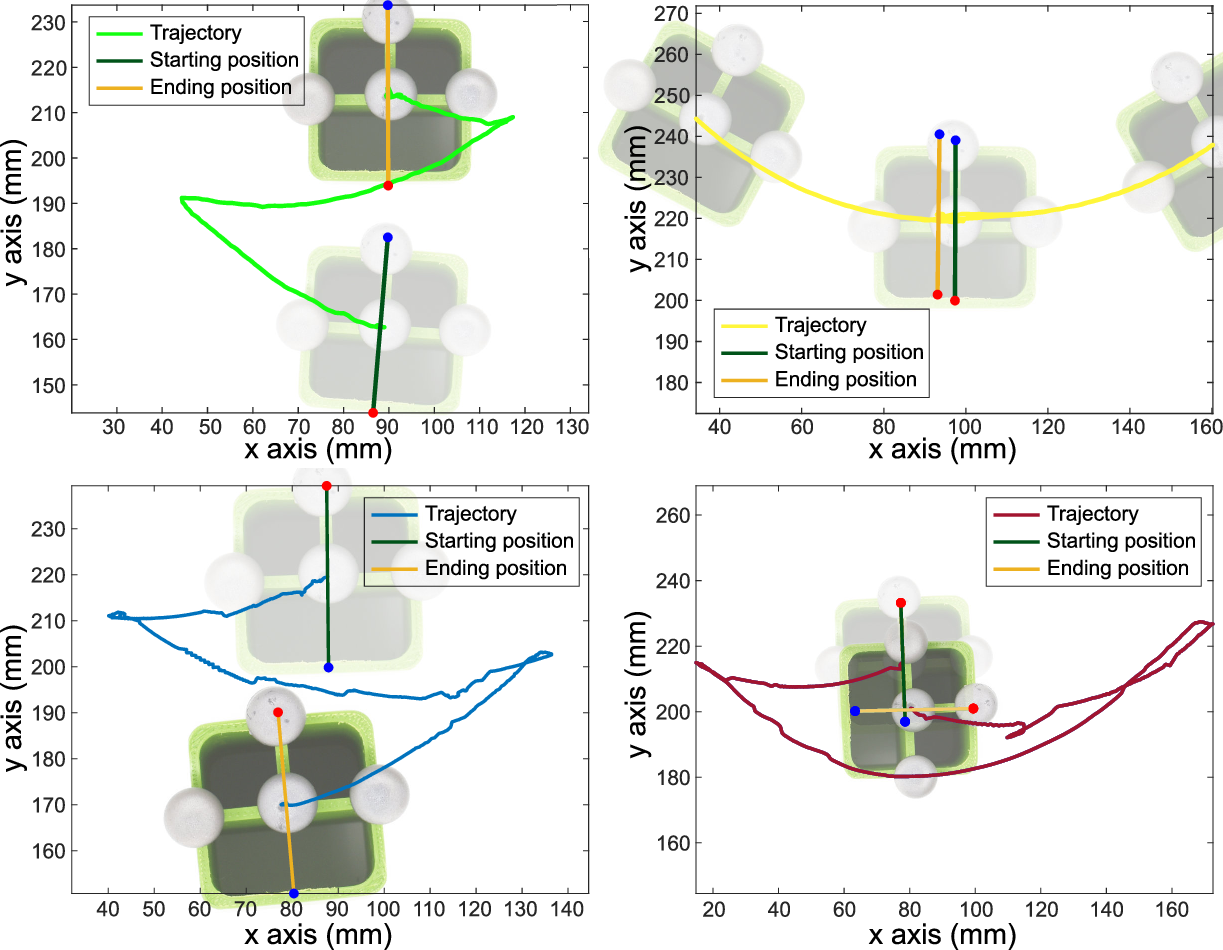}
    \caption{Trajectories of a 40 mm square manipulated with normal and O-VF surface (medium density). \textbf{Top:} Low friction normal surface trajectory (left) and high friction normal surface trajectory (right). \textbf{Bottom:} Translational trajectory (left) and rotational trajectory (right) of O-VF surface. Red and blue dots indicate the two markers on the object. The green and yellow lines indicate the object starting and ending position, respectively.}
    \label{trajectory}
\end{figure}

\begin{figure*}[t!]
    \centering
    \includegraphics[width=\textwidth]{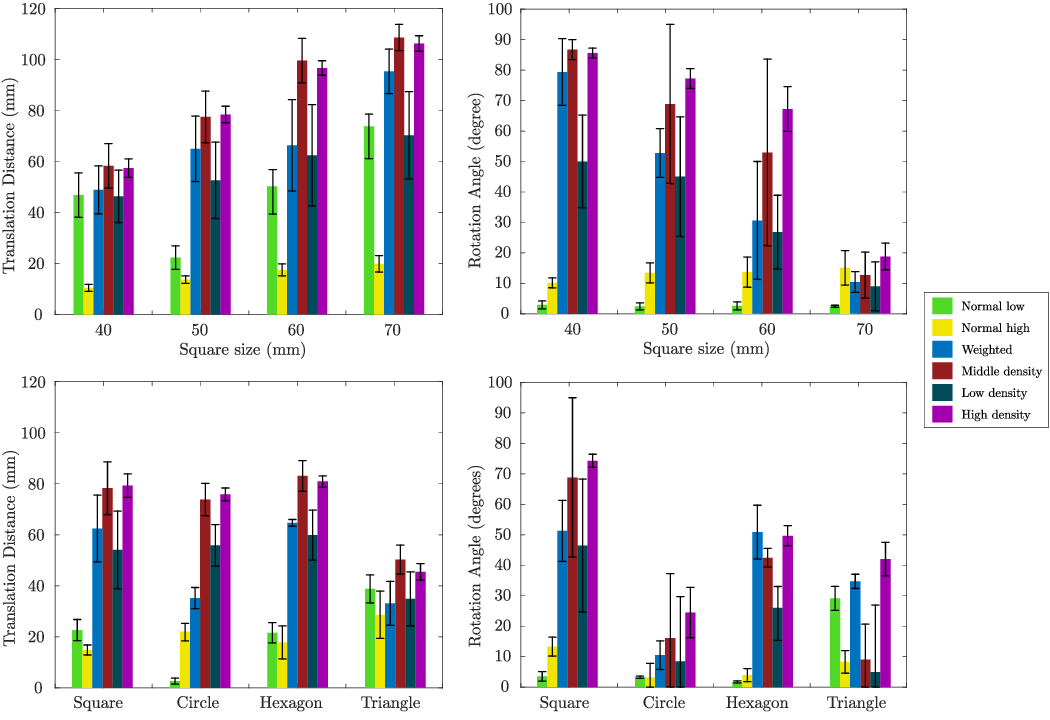}
    \caption{Graphs showing the variation in translation and rotation capability for varying shapes and sizes for the six tested finger surfaces (low friction normal, high friction normal, weighted O-VF, medium density O-VF, low density O-VF, and high density O-VF), with standard deviation for each bar shown. Upper two graphs show the translational and rotational capability on 4 sizes of squares, while the lower two graphs show the same capability on 4 alternative shapes.}
    \label{Results}
\end{figure*}

Fig.~\ref{trajectory} shows the translation trajectory of the 40 mm square manipulated with the constant friction and O-VF middle density surface under the same control algorithm of the servo motors.  Fig.~\ref{Results} shows the variation of translation and rotation capabilities for varying square sizes and shapes with these 6 finger surfaces. In this test, we set the performance of constant friction surface as the baseline results. We then compare the O-VF surfaces performance to the baseline to observe the manipulation improvement. Additionally, we compare the performance between O-VF surfaces to see the effect of different design parameters. 

In Fig.~\ref{Results}, the green and yellow lines are the manipulation results for constant low friction and high friction surfaces. The translation distance of various square sizes are similar for the high friction surface, but for the low friction surface the performance varies for different object sizes, which is related to the nominal gap between the two fingers on the hand design. The nominal gap of this hand design is around 50 mm. When grasping smaller objects ($<$50 mm), the fingers rotate pointing towards to each other instead of staying parallel, which means there are gaps between the fingers and the objects at the starting position. Once the finger rotates with the objects, the hand object system become as a slider-crank, the object slides towards to the finger base. On the contrary, for larger objects ($>$50 mm) the objects slide towards the fingertips. This also explains why the translation distance of the 50 mm square is significantly smaller than others, as the direction of object translation is not affected by the fingers. For the constant high friction surface, the objects are more likely to rotate between the nominal gap of the hand instead of sliding. From Fig.~\ref{trajectory} (top right) the object rotates along the fingers during the manipulation, but returns to its original position roughly at the end of the manipulation. Further, both translation and rotation for constant friction surfaces are not controllable.

The overall performance of the O-VF surfaces on square objects are better than the constant friction surfaces (Fig.~\ref{Results}). The gripper with active O-VF surface fingers had a significantly larger rotation angle, ranging from a mean of 13.46 mm to 83.27 mm, compare to the normal constant friction surface gripper, whose range is from a mean of 6.18 mm to 8.24 mm. The size of the object has a significant influence on the rotation capability of the O-VF surfaces, with larger objects showing a decrease in rotation. For the 70 mm square, the O-VF surfaces cannot significantly improve the rotation better than the constant friction surfaces due to the nominal gap of the hand design is only around 50 mm.

In contrast, the translation distances show no obvious trend for each of the different shapes. However, the translation differences between the O-VF surface and constant friction surface are significant, especially on square, circle, and hexagon. For the constant friction surface, the average translation distances of the square, circle, hexagon, and triangle were 19.8, 12.6, 20.3, and 38.6 mm. The results show the O-VF surface increases the translation ability of the gripper greatly, apart from the triangle object. This may be because the contact type for the triangular object is different from the other shapes, as it cannot achieve surface (planar) contact for both fingers at the same time, achieving instead one finger with pivot contact and the other with planar contact. When the finger object contact model is pivot at a high friction surface, the object will rotate, or must wait for the finger pad to rotate to instead generate a planar contact model to perform sliding. Therefore in this case, for the same control scheme, fingers need to rotate more to achieve the same amount of translation.

Fig.~\ref{Results}(bottom right) shows the rotation angles for different objects. For most of the O-VF surface design, the rotational capability of circle and triangle are limited, with the circle showing almost no overall rotation (less than 15$^{\circ}$), while the hexagon showed a greater value (average of 43.25$^{\circ}$) but still smaller compared to that of the square (60.74$^{\circ}$). With the constant friction surface, the gripper showed consistent performance across shapes, however the rotation angles were minimal with a mean of 14.84$^{\circ}$ for the square, 3.16$^{\circ}$ for the circle, 3.89$^{\circ}$ for the hexagon, and 20.21$^{\circ}$ for the triangle.

Some of the rotation results show large standard deviations (more than 20$^{\circ}$), indicating the rotation capability of the O-VF surfaces with certain design parameters are not stable under the same open-loop control approach. This can be explained due to the object contact model with the O-VF surface. If the object shape and the finger pad are more likely to generate pivot contacts with the low friction surface, the rotation of the object will be harder to control, as objects will have self-alignment (rotation) generating a stable contact model (in this work, a planar contact model). In Fig.~\ref{Results}(bottom right), the results show that the active O-VF surface works well on square and hexagon shapes, which both can have planar contact with both fingers, whereas the circle shows almost no change in rotation as it is always in point contact with both fingers, therefore making sliding motion hard to control.

By adjusting the design parameters of the O-VF surfaces, a variation in the performance of the gripper was achieved. Overall, the weighted high/low friction length design performed worse than the equally weighted surfaces. With a weighted design, the object has higher possibility of getting stuck in the gaps while sliding over the low friction surface. On the other hand, with the higher ratio on high friction length, the design performed better on rotating various shapes. Unit density is another factor we evaluated in the experiments. According to the mathematics model, higher density has lower overall structure height change ($\Delta h$), and smaller valley gaps. Experimental results showed that the $\Delta h$ effected the in-hand manipulation capability a lot. Higher density showed better manipulation capability and was more stable. Although the mean value of the translation distance and rotation angle of the middle and high density design are similar, the standard deviation of the high density are much smaller, indicating a higher reliability. The low density O-VF surface in comparison performed worse, showing similar performance to the weighted design.

In comparison to the hand developed by Spiers \textit{et al.} \cite{spiers2018variable}, our surface design achieved the same rotation per cycle ($\sim$90\textdegree) in a more condensed form. As our gripper and finger size are different, the values of the translation are incomparable, as they are limited by the length of the finger and range of motion of the fingers, not the surface. Additionally, they did not evaluate shapes such as the triangle or hexagon. Further, we have provided additional experimentation on objects larger than the gripper nominal gap. Most importantly, our design is parametric and can be optimised to provide enhanced manipulation if the object sizes are known.

\section{Conclusion}
This work proposes and evaluates a novel origami-inspired variable friction (O-VF) surface design, producing a simple two-fingered two-degree-of-freedom robotic gripper capable of achieving translation and rotation object manipulation. The proposed O-VF surface design is parametric, and the design parameters and material selection were explored considering the effect on in-hand manipulation. The design is also capable of matching requirements in terms of the size and thickness of the surface, the ratio of the high to low friction contact areas, the change in structure height after folding, and the coefficient of friction of the high and low friction surfaces. Using an open loop control approach, 6 finger surfaces (2 constant, 4 O-VF) were evaluated in terms of translation and rotation. Results show the unit density is one of the main aspects observed to improve the gripper performance, showing a higher manipulation magnitude per cycle with a higher reliability. For the objects manipulated, it was also observed that objects with faces parallel to each of the fingers produced a larger manipulation as the contact friction could be varied, unlike in point contact. For future work, the performance of the O-VF surface could be improved to move an object to a target position and orientation via closed-loop control, with the addition of vision or tactile sensors to monitor translation and rotation magnitude. In addition, the size of the O-VF surface could be scaled down while increasing the unit density to manufacture an origami soft skin.

\addtolength{\textheight}{-0cm}   

\bibliographystyle{IEEEtran}
\bibliography{references.bib}

\begin{thebibliography}{10}
\providecommand{\url}[1]{#1}
\csname url@samestyle\endcsname
\providecommand{\newblock}{\relax}
\providecommand{\bibinfo}[2]{#2}
\providecommand{\BIBentrySTDinterwordspacing}{\spaceskip=0pt\relax}
\providecommand{\BIBentryALTinterwordstretchfactor}{4}
\providecommand{\BIBentryALTinterwordspacing}{\spaceskip=\fontdimen2\font plus
\BIBentryALTinterwordstretchfactor\fontdimen3\font minus
  \fontdimen4\font\relax}
\providecommand{\BIBforeignlanguage}[2]{{%
\expandafter\ifx\csname l@#1\endcsname\relax
\typeout{** WARNING: IEEEtran.bst: No hyphenation pattern has been}%
\typeout{** loaded for the language `#1'. Using the pattern for}%
\typeout{** the default language instead.}%
\else
\language=\csname l@#1\endcsname
\fi
#2}}
\providecommand{\BIBdecl}{\relax}
\BIBdecl

\bibitem{guo2017design}
M.~Guo, D.~V. Gealy, J.~Liang, J.~Mahler, A.~Goncalves, S.~McKinley, J.~A.
  Ojea, and K.~Goldberg, ``Design of parallel-jaw gripper tip surfaces for
  robust grasping,'' in \emph{IEEE International Conference on Robotics and
  Automation (ICRA)}, 2017, pp. 2831--2838.

\bibitem{ueda2010multifingered}
J.~Ueda, M.~Kondo, and T.~Ogasawara, ``The multifingered naist hand system for
  robot in-hand manipulation,'' \emph{Mechanism and Machine Theory}, vol.~45,
  no.~2, pp. 224--238, 2010.

\bibitem{chella2004posture}
A.~Chella, H.~D{\v{z}}indo, I.~Infantino, and I.~Macaluso, ``A posture sequence
  learning system for an anthropomorphic robotic hand,'' \emph{Robotics and
  Autonomous Systems}, vol.~47, no. 2-3, pp. 143--152, 2004.

\bibitem{shi2017dynamic}
J.~Shi, J.~Z. Woodruff, P.~B. Umbanhowar, and K.~M. Lynch, ``Dynamic in-hand
  sliding manipulation,'' \emph{IEEE Transactions on Robotics}, vol.~33, no.~4,
  pp. 778--795, 2017.

\bibitem{rojas2016gr2}
N.~Rojas, R.~R. Ma, and A.~M. Dollar, ``The gr2 gripper: an underactuated hand
  for open-loop in-hand planar manipulation,'' \emph{IEEE Transactions on
  Robotics}, vol.~32, no.~3, pp. 763--770, 2016.

\bibitem{bircher2017two}
W.~G. Bircher, A.~M. Dollar, and N.~Rojas, ``A two-fingered robot gripper with
  large object reorientation range,'' in \emph{IEEE International Conference on
  Robotics and Automation (ICRA)}, 2017, pp. 3453--3460.

\bibitem{dafle2014extrinsic}
N.~C. Dafle, A.~Rodriguez, R.~Paolini, B.~Tang, S.~S. Srinivasa, M.~Erdmann,
  M.~T. Mason, I.~Lundberg, H.~Staab, and T.~Fuhlbrigge, ``Extrinsic dexterity:
  In-hand manipulation with external forces,'' in \emph{IEEE International
  Conference on Robotics and Automation (ICRA)}, 2014, pp. 1578--1585.

\bibitem{chavan2015two}
N.~Chavan-Dafle, M.~T. Mason, H.~Staab, G.~Rossano, and A.~Rodriguez, ``A
  two-phase gripper to reorient and grasp,'' in \emph{IEEE International
  Conference on Automation Science and Engineering (CASE)}, 2015, pp.
  1249--1255.

\bibitem{chavan2018pneumatic}
N.~Chavan-Dafie, K.~Lee, and A.~Rodriguez, ``Pneumatic shape-shifting fingers
  to reorient and grasp,'' in \emph{IEEE 14th International Conference on
  Automation Science and Engineering (CASE)}, 2018, pp. 988--993.

\bibitem{ward2017model}
B.~Ward-Cherrier, N.~Rojas, and N.~F. Lepora, ``Model-free precise in-hand
  manipulation with a 3d-printed tactile gripper,'' \emph{IEEE Robotics and
  Automation Letters}, vol.~2, no.~4, pp. 2056--2063, 2017.

\bibitem{terasaki1994motion}
H.~Terasaki and T.~Hasegawa, ``Motion planning for intelligent manipulations by
  sliding and rotating operations with parallel two-fingered grippers,'' in
  \emph{IEEE/RSJ International Conference on Intelligent Robots and Systems
  (IROS)}, vol.~1, 1994, pp. 119--126.

\bibitem{chorley2008biologically}
C.~Chorley, C.~Melhuish, T.~Pipe, J.~Rossiter, and G.~Whiteley, ``A
  biologically inspired fingertip design for compliance and strength,''
  \emph{Proceeding of Towards Autonomous Robotic Systems}, pp. 239--244, 2008.

\bibitem{lu2019soft}
Q.~Lu and N.~Rojas, ``On soft fingertips for in-hand manipulation: Modeling and
  implications for robot hand design,'' \emph{IEEE Robotics and Automation
  Letters}, vol.~4, no.~3, pp. 2471--2478, 2019.

\bibitem{tomlinson2007review}
S.~Tomlinson, R.~Lewis, and M.~Carr{\'e}, ``Review of the frictional properties
  of finger-object contact when gripping,'' \emph{Proceedings of the
  Institution of Mechanical Engineers, Part J: Journal of Engineering
  Tribology}, vol. 221, no.~8, pp. 841--850, 2007.

\bibitem{comaish1971skin}
S.~Comaish and E.~Bottoms, ``The skin and friction: deviations from amonton's
  laws, and the effects of hydration and lubrication,'' \emph{British Journal
  of Dermatology}, vol.~84, no.~1, pp. 37--43, 1971.

\bibitem{adams2013finger}
M.~J. Adams, S.~A. Johnson, P.~Lef{\`e}vre, V.~L{\'e}vesque, V.~Hayward,
  T.~Andr{\'e}, and J.-L. Thonnard, ``Finger pad friction and its role in grip
  and touch,'' \emph{Journal of The Royal Society Interface}, vol.~10, no.~80,
  p. 20120467, 2013.

\bibitem{spiers2018variable}
A.~J. Spiers, B.~Calli, and A.~M. Dollar, ``Variable-friction finger surfaces
  to enable within-hand manipulation via gripping and sliding,'' \emph{IEEE
  Robotics and Automation Letters}, vol.~3, no.~4, pp. 4116--4123, 2018.

\bibitem{rus2018design}
D.~Rus and M.~T. Tolley, ``Design, fabrication and control of origami robots,''
  \emph{Nature Reviews Materials}, vol.~3, no.~6, p. 101, 2018.

\bibitem{miura1985method}
K.~Miura, ``Method of packaging and deployment of large membranes in space,''
  \emph{The Institute of Space and Astronautical Science Report}, vol. 618,
  p.~1, 1985.

\bibitem{kresling2012origami}
B.~Kresling, ``Origami-structures in nature: lessons in designing “smart”
  materials,'' \emph{MRS Online Proceedings Library Archive}, vol. 1420, 2012.

\bibitem{ultmat}
\BIBentryALTinterwordspacing
Ultimaker, \emph{Material Technical and Safety Data Sheets}, 2018, version
  4.002. [Online]. Available:
  \url{https://ultimaker.com/en/resources/50461-technical-and-safety-data-sheets}
\BIBentrySTDinterwordspacing

\bibitem{ma2017yale}
R.~Ma and A.~Dollar, ``Yale openhand project: Optimizing open-source hand
  designs for ease of fabrication and adoption,'' \emph{IEEE Robotics \&
  Automation Magazine}, vol.~24, no.~1, pp. 32--40, 2017.

\end{thebibliography}

\end{document}